\documentclass[times,twocolumn,final]{elsarticle}
\usepackage{medima}
\usepackage{framed,multirow}
\usepackage{amssymb}
\usepackage{latexsym}
\usepackage{url}
\usepackage{float}
\usepackage{hyperref}
\usepackage{multirow}
\usepackage{amsmath}
\usepackage{booktabs}
\usepackage{placeins}
\usepackage{subcaption}
\usepackage[dvipsnames]{xcolor}
\usepackage{graphicx}
\usepackage{adjustbox}

\definecolor{newcolor}{rgb}{.8,.349,.1}

\journal{Expert Systems with Applications}

\begin{document}

\verso{WaveRNet: Wavelet-Guided Frequency Learning for Domain-Generalized Retinal Vessel Segmentation}

\begin{frontmatter}

\title{WaveRNet: Wavelet-Guided Frequency Learning for Multi-Source Domain-Generalized Retinal Vessel Segmentation}

\author[1]{Chanchan \snm{Wang}}
\ead{wusheng070@gmail.com}
\author[2]{Yuanfang \snm{Wang}}
\ead{outlook_A245B73066013466@outlook.com}
\author[3]{Qing \snm{Xu}}
\ead{scxqx1@nottingham.ac.uk}
\author[1]{Guanxin \snm{Chen}\corref{cor1}}
\ead{blacknut@xju.edu.cn}

\cortext[cor1]{Corresponding author.} 

\address[1]{School of Software, Xinjiang University, Urumqi, Xinjiang, China.}
\address[2]{The First Clinical Medical College, Nanjing University of Chinese Medicine, Nanjing, China.}
\address[3]{School of Computer Science, University of Nottingham, UK.}

\begin{abstract}
Domain-generalized retinal vessel segmentation is critical for automated ophthalmic diagnosis, yet faces significant challenges from domain shift induced by non-uniform illumination and varying contrast, compounded by the difficulty of preserving fine vessel structures. While the Segment Anything Model (SAM) exhibits remarkable zero-shot capabilities, existing SAM-based methods rely on simple adapter fine-tuning while overlooking frequency-domain information that encodes domain-invariant features, resulting in degraded generalization under illumination and contrast variations. Furthermore, SAM's direct upsampling inevitably loses fine vessel details. To address these limitations, we propose WaveRNet, a wavelet-guided frequency learning framework for robust multi-source domain-generalized retinal vessel segmentation. Specifically, we devise a Spectral-guided Domain Modulator (SDM) that integrates wavelet decomposition with learnable domain tokens, enabling the separation of illumination-robust low-frequency structures from high-frequency vessel boundaries while facilitating domain-specific feature generation. Furthermore, we introduce a Frequency-Adaptive Domain Fusion (FADF) module that performs intelligent test-time domain selection through wavelet-based frequency similarity and soft-weighted fusion. Finally, we present a Hierarchical Mask-Prompt Refiner (HMPR) that overcomes SAM's upsampling limitation through coarse-to-fine refinement with long-range dependency modeling. Extensive experiments under the Leave-One-Domain-Out protocol on four public retinal datasets demonstrate that WaveRNet achieves state-of-the-art generalization performance. The source code is available at \url{https://github.com/Chanchan-Wang/WaveRNet}.
\end{abstract}


\end{frontmatter}

%\linenumbers

\section{Introduction}
\label{sec:intro}

Retinal photography constitutes a cornerstone of ophthalmic imaging, with devices from manufacturers including Canon, Topcon, and Zeiss deployed across diverse clinical settings worldwide. Retinal vessel segmentation from retinal images is essential for diagnosing and monitoring ophthalmic diseases including diabetic retinopathy, glaucoma, and hypertension \citep{qin2024review,liu2025deep}. Accurate vessel delineation enables quantitative analysis of vascular morphology, such as tortuosity, branching patterns, and arteriovenous ratio, which are critical biomarkers for disease assessment \citep{fraz2012blood,l2017recent}. However, building a universal vessel segmentation model that generalizes well to unseen imaging domains remains challenging. The primary obstacle is domain shift caused by non-uniform illumination and contrast variations across different devices and clinical sites \citep{liao2024dual,ye2026advancing}. Specifically, retinal images typically exhibit a bright central region with darker peripheral areas due to uneven light distribution, and contrast levels vary substantially depending on imaging protocols and patient-specific factors such as retinal pigmentation \citep{li2025retinal,kumar2024optimal}. Additionally, retinal vessels present inherent segmentation difficulties due to fine vessel structures that are easily lost during feature extraction \citep{chen2021retinal,ding2024rcar}.

Deep learning methods built upon U-Net \citep{ronneberger2015u} and its variants \citep{jin2019dunet,luo2025pa,qin2025medical} have demonstrated remarkable effectiveness on single-dataset benchmarks through hierarchical spatial feature learning. However, these models assume that training and testing data follow the same distribution, leading to significant performance degradation when applied to images from different devices or clinical sites \citep{hu2024domain}. This limitation highlights the need for domain-generalized approaches that can explicitly handle illumination and contrast variations inherent in multi-source retinal imaging.

Diverse domain generalization strategies have emerged for medical image segmentation. Data augmentation methods apply random intensity and color transformations to simulate appearance variations during training \citep{huang2025generative,wen2024denoising}. Feature alignment approaches employ adversarial learning or moment matching to extract domain-invariant representations \citep{xu2025aegis,zhang2025adverin}. Frequency-based methods leverage Fourier transform to separate low-frequency content from high-frequency style information for domain adaptation \citep{zhang2025domain,liu2024wavelet}. However, these approaches have notable limitations for retinal vessel segmentation. Data augmentation cannot comprehensively cover the complex illumination patterns in real clinical images. Feature alignment methods often sacrifice fine-grained discriminability for domain invariance, impairing the detection of tiny capillaries. Existing frequency-based methods primarily target style transfer rather than explicitly addressing illumination and contrast variations, which are the primary causes of domain shift in retinal imaging \citep{galappaththige2024generalizing}.

The Segment Anything Model (SAM) \citep{kirillov2023segment} represents a paradigm shift in visual segmentation, leveraging over one billion mask annotations to achieve unprecedented zero-shot generalization across diverse visual domains. This capability has motivated its adaptation to medical imaging, where domain shift is prevalent. MedSAM \citep{ma2024segment} fine-tunes SAM on large-scale medical datasets, and subsequent works have shown promising results across various medical segmentation tasks \citep{mazurowski2023segment,zhang2024segment,gao2024desam}. However, existing SAM-based approaches share two critical limitations. First, they perform adaptation exclusively in the spatial feature domain while neglecting frequency-domain information. For retinal vessel segmentation, this oversight is particularly problematic because non-uniform illumination primarily affects low-frequency components, while contrast variations impact high-frequency edge information \citep{vasu2025optimal}. By ignoring this frequency perspective, current methods fail to address the root causes of domain shift in retinal imaging. Second, the original SAM decoder upsamples low-resolution features directly to high-resolution masks through simple transposed convolutions. This abrupt resolution jump inevitably loses fine vessel details, where tiny capillaries and delicate branching structures require gradual, fine-grained reconstruction, causing vessel discontinuities and boundary inaccuracies \citep{zhu2025duws}.

To address these limitations, we propose WaveRNet, a wavelet-guided frequency learning framework for robust multi-source domain-generalized retinal vessel segmentation. Our core insight is that wavelet transform can effectively decompose features into frequency components, enabling explicit modeling of illumination-robust and contrast-aware representations. Specifically, we devise a Spectral-guided Domain Modulator (SDM) that integrates wavelet decomposition with domain-specific modulation. SDM employs learnable dual-branch convolutions to separate low-frequency components encoding illumination-stable global structures from high-frequency components capturing contrast-sensitive vessel boundaries, while learnable domain tokens enable discriminative feature generation tailored to each source domain's imaging characteristics. Furthermore, we introduce a Frequency-Adaptive Domain Fusion (FADF) module for intelligent test-time inference, which computes wavelet-based frequency similarity between test images and source domains, performing soft-weighted fusion to handle unseen domains. Finally, we present a Hierarchical Mask-Prompt Refiner (HMPR) that overcomes SAM's direct upsampling limitation through progressive coarse-to-fine refinement, where each stage's output serves as the mask prompt for subsequent refinement, enhanced by attention mechanisms for long-range dependency modeling. The contributions of this work are summarized as follows:

\begin{itemize}
    \item We devise SDM that integrates wavelet transform with learnable domain tokens, decomposing features into low-frequency and contrast-sensitive high-frequency components for domain-specific feature generation.

    \item We introduce FADF for intelligent test-time domain adaptation via wavelet-based frequency similarity and soft-weighted fusion across source domains.

    \item We design HMPR to address SAM's upsampling limitation through coarse-to-fine mask-prompt refinement with long-range dependency modeling, progressively recovering fine vessel details.

    \item Extensive experiments under the Leave-One-Domain-Out protocol on four public retinal datasets demonstrate that WaveRNet achieves superior generalization capabilities.
\end{itemize}

\section{Related Work}
\label{sec:related}

\subsection{Retinal Vessel Segmentation}

The encoder-decoder paradigm, exemplified by U-Net \citep{ronneberger2015u}, has established a foundational architecture for medical image segmentation through its elegant skip connection design. The past decade has witnessed substantial research efforts toward enhancing feature extraction capabilities for retinal vessel segmentation \citep{fraz2012blood,chen2021retinal}. Early CNN-based approaches exploited inductive biases to capture local spatial patterns, while Transformer-based architectures subsequently expanded model capacity through self-attention mechanisms that capture long-range dependencies \citep{qin2024review,liu2025deep}.

Building upon this foundation, numerous methodologies have emerged to enhance U-Net for retinal vessel segmentation from complementary perspectives. Dense skip connections and attention mechanisms have been widely adopted to capture multi-scale features and emphasize vessel-relevant regions \citep{zhou2018unetpp,oktay2018attention,xu2023dcsau}. Residual learning has been integrated to enable deeper networks for fine vessel detection \citep{jha2019resunetpp,ibtehaz2020multiresunet,ibtehaz2023accunet}. Deformable convolutions have been employed to adapt to tortuous vessel structures \citep{jin2019dunet}. Lightweight architectures have been developed for efficient inference while maintaining segmentation accuracy \citep{valanarasu2022unext,dinh20231m}. Multi-scale feature fusion strategies have been proposed to handle vessels of varying widths \citep{wu2024mfmsnet}. Transformer-based methods have modeled long-range dependencies through self-attention for improved global context understanding \citep{cao2022swin,rahman2024emcad}. Despite these advancements, existing methods are typically trained under consistent imaging conditions and suffer performance degradation when deployed across different clinical sites due to non-uniform illumination and contrast variations. Moreover, these approaches neglect how domain-specific variations are encoded in different frequency bands. Unlike existing methods, our work explicitly leverages wavelet-based frequency decomposition to disentangle illumination-robust and contrast-sensitive features, enabling the learning of domain-invariant representations.

\subsection{SAM for Medical Image Segmentation}

SAM \citep{kirillov2023segment} has established itself as a transformative foundation model for image segmentation, trained on an unprecedented scale of over one billion masks spanning diverse natural images. SAM demonstrates impressive zero-shot generalization through its prompt-based paradigm, with an architecture consisting of a Vision Transformer image encoder, a flexible prompt encoder, and a lightweight mask decoder. Recently, SAM2 \citep{ravi2024sam} has been released with improved efficiency. The strong generalization ability has motivated SAM's adaptation to medical imaging \citep{xu2025hrmedseg, xu2025lightsam}. MedSAM \citep{ma2024segment} fine-tunes SAM on large-scale medical datasets spanning multiple modalities. SAM-Med2D \citep{cheng2023sammed2d} introduces medical-specific adaptations for clinical images. Various adapter-based methods have been proposed to efficiently transfer SAM's knowledge to specific medical domains \citep{zhang2024segment,gao2024desam}.

Several studies have explored SAM-based approaches for ophthalmic image segmentation. \citet{fazekas2023adapting} conducted a comprehensive evaluation of SAM adaptations on retinal OCT fluid segmentation, demonstrating SAM's potential in retinal imaging through adapter-based fine-tuning. \citet{qiu2023learnable} proposed Learnable Ophthalmology SAM with learnable prompt layers to adapt SAM for multi-modal ophthalmic images, including retinal vessel and OCT layer segmentation. While these methods benefit from SAM's pretrained representations, they share two common limitations. First, the original decoder directly upsamples low-resolution features to high-resolution masks through simple transposed convolutions, causing fine capillaries and delicate branching structures to be easily lost during the upsampling process. Second, existing methods typically rely on a single domain during inference, lacking adaptive mechanisms to leverage multi-domain knowledge for unseen test samples. While these methods benefit from SAM's pretrained representations, the original decoder directly upsamples low-resolution features to high-resolution masks, causing fine capillaries and delicate branching structures to be lost during upsampling. Moreover, existing methods lack adaptive mechanisms to leverage multi-domain knowledge for unseen test samples. Different from these approaches, we introduce frequency-aware modulation and progressive refinement mechanisms to address both the fine structure preservation and cross-domain generalization challenges.

\section{Method}
\label{sec:method}

\subsection{Overview}

Given retinal images from $K$ source domains $\mathcal{D} = \{D_1, D_2, ..., D_K\}$, domain-generalized vessel segmentation aims to train a model on labeled source domains and generalize to an unseen target domain $D_t$ without accessing any target data during training. Each source domain $D_k = \{(x_i^k, y_i^k)\}_{i=1}^{N_k}$ contains $N_k$ retinal images $x_i^k$ with corresponding vessel annotations $y_i^k$. The objective is to learn domain-invariant representations that capture vessel structures while being robust to illumination and contrast variations across domains.

\begin{figure*}[!t]
\centering
\includegraphics[width=\textwidth]{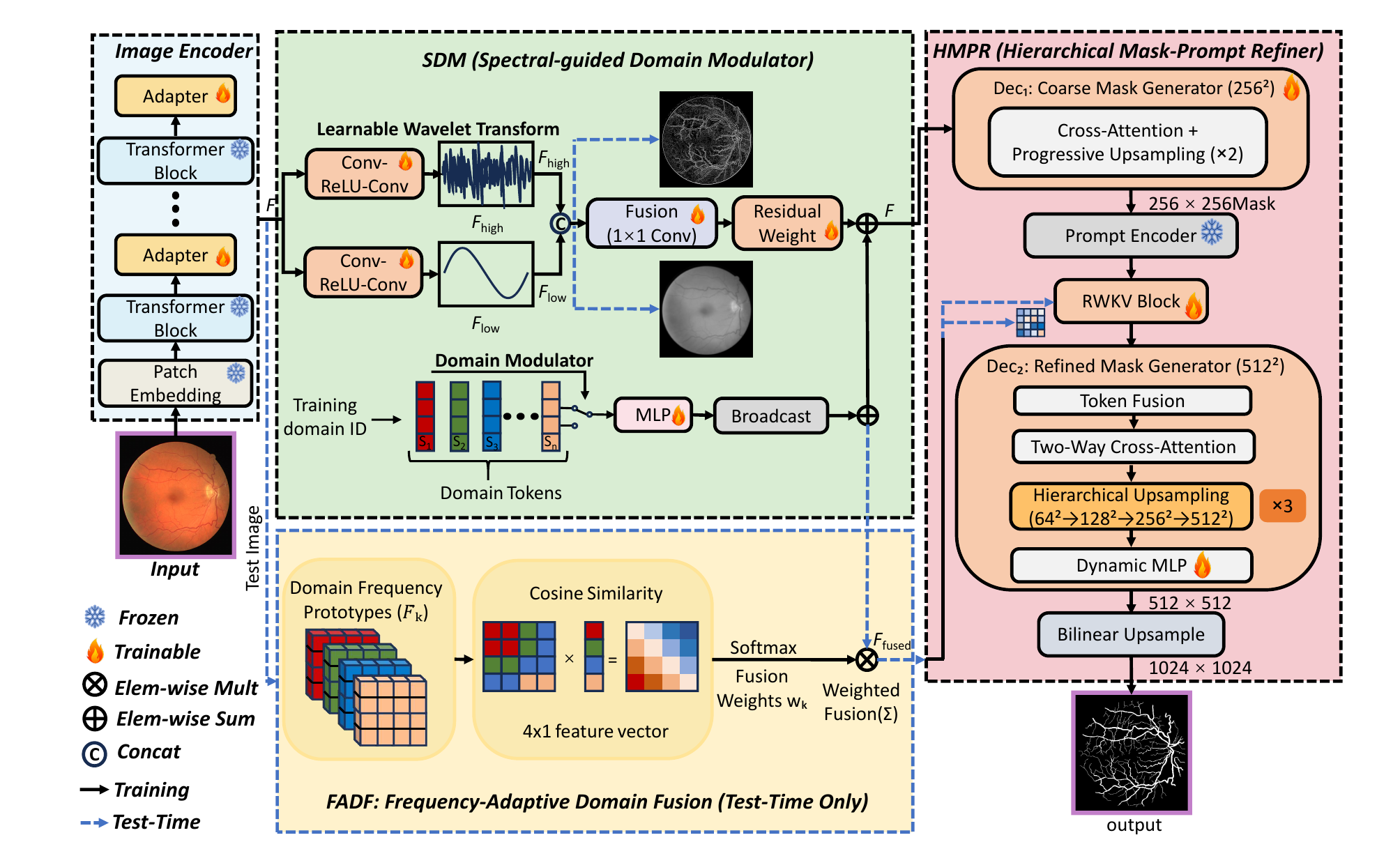}
\caption{Overview of the proposed WaveRNet framework. The Spectral-guided Domain Modulator (SDM) decomposes features into high-frequency and low-frequency components through learnable wavelet transform, followed by domain-specific modulation using learnable domain tokens. During inference, Frequency-Adaptive Domain Fusion (FADF) computes wavelet-based frequency similarity for intelligent test-time domain selection. The Hierarchical Mask-Prompt Refiner (HMPR) progressively refines segmentation through coarse-to-fine mask generation with cross-attention and hierarchical upsampling.}
\label{fig:framework}
\end{figure*}

The overall architecture of WaveRNet is illustrated in Fig.~\ref{fig:framework}. Given an input retinal image $x \in \mathbb{R}^{H \times W \times 3}$, we first extract image embeddings through the SAM image encoder with adapter layers: $F \in \mathbb{R}^{C \times H' \times W'}$, where $H' = H/16$ and $W' = W/16$. The extracted features are then processed by our SDM, which decomposes features into high-frequency and low-frequency components through learnable wavelet transform, followed by domain-specific modulation using learnable domain tokens. During inference, FADF computes wavelet-based frequency similarity between the test image and source domains, performing soft-weighted fusion to generate domain-adaptive features. Finally, the HMPR progressively refines the segmentation through a coarse-to-fine strategy, where each stage's prediction serves as the mask prompt for the subsequent stage. The final vessel segmentation mask $\hat{y} \in \mathbb{R}^{H \times W}$ is obtained after the refinement process. Through the synergy of frequency-aware domain adaptation and hierarchical mask refinement, WaveRNet achieves robust cross-domain generalization while preserving fine vessel structures. This unified framework coherently addresses the two fundamental challenges in domain-generalized retinal vessel segmentation: the frequency-aware modulation in SDM and FADF mitigates domain shift induced by illumination and contrast variations, while the progressive refinement in HMPR recovers fine vessel structures that would otherwise be lost during direct upsampling.

\subsection{Spectral-guided Domain Modulator}

Existing SAM-based methods \citep{cheng2023sammed2d, gao2024desam} adapt to medical imaging solely in the spatial feature domain, neglecting frequency-domain information that is inherently robust to cross-domain variations. In retinal imaging, non-uniform illumination primarily manifests in low-frequency components, while contrast variations affect high-frequency edge information. To explicitly model these frequency characteristics, SDM integrates wavelet transform with domain-specific modulation. Traditional discrete wavelet transform (DWT) decomposes a signal into approximation coefficients (low-frequency) and detail coefficients (high-frequency) through predefined filter banks. Given a 2D feature map $F \in \mathbb{R}^{C \times H \times W}$, the standard DWT applies low-pass filter $\phi$ and high-pass filter $\psi$ along rows and columns:
\begin{equation}
    F_{\rm LL} = (\phi * (\phi * F)^T)^T, \quad F_{\rm HH} = (\psi * (\psi * F)^T)^T
\end{equation}
where $F_{\rm HH}$ captures high-frequency edge details. However, predefined wavelet filters may not optimally separate domain-invariant structures from domain-specific variations in retinal images. Motivated by the observation that task-specific frequency characteristics require adaptive decomposition, we propose a learnable wavelet transform that employs dual-branch convolutions to adaptively decompose features:
\begin{equation}
    F_{\rm low} =W_{\rm low}(F) = \sigma(\text{Conv}_{3\times3}(\sigma(\text{Conv}_{3\times3}(F))))
\end{equation}
\begin{equation}
    F_{\rm high} = W_{\rm high}(F) = \sigma(\text{Conv}_{3\times3}(\sigma(\text{Conv}_{3\times3}(F))))
\end{equation}
where $\sigma$ denotes the ReLU activation, and $W_{\rm low}$ and $W_{\rm high}$ are implemented as separate convolutional branches that learn to extract low-frequency illumination-stable structures and high-frequency contrast-sensitive boundaries, respectively.

The decomposed components are fused through a $1\times1$ convolution to generate frequency-aware features:
\begin{equation}
    F_{\rm wave} = \text{Conv}_{1\times1}([F_{\rm low}; F_{\rm high}]) + \alpha \cdot F
\end{equation}
where $[\cdot;\cdot]$ denotes channel-wise concatenation, and $\alpha$ is a learnable residual weight initialized to a small value to ensure stable training. To capture the unique imaging characteristics of each source domain, we introduce a Domain Modulator with learnable domain tokens. For $K$ source domains, we maintain a set of domain tokens $\{t_k\}_{k=1}^{K}$, where $t_k \in \mathbb{R}^{C}$ encodes domain-specific information. Each domain token is associated with a lightweight MLP network:
\begin{equation}
    \hat{t}_k = \text{MLP}_k(t_k) = W_2^k \cdot \sigma(W_1^k \cdot t_k)
\end{equation}
where $W_1^k \in \mathbb{R}^{C/4 \times C}$ and $W_2^k \in \mathbb{R}^{C \times C/4}$ are domain-specific projection matrices. During training, given an image from domain $k$, the modulated token is spatially broadcast and added to the wavelet-enhanced features:
\begin{equation}
    F_{\rm SDM} = F_{\rm wave} + \mathrm{broadcast}(\hat{t}_k)
\end{equation}
where $\mathrm{broadcast}(\cdot): \mathbb{R}^{C} \rightarrow \mathbb{R}^{C \times H \times W}$ replicates the channel-wise token across spatial dimensions. This design enables SDM to generate discriminative features tailored to each domain's illumination and contrast characteristics while preserving the frequency-aware representations learned through wavelet decomposition.

\subsection{Frequency-Adaptive Domain Fusion}

During inference, the domain identity of test images is unknown, making it challenging to select the appropriate domain token. Existing methods either use a single shared adapter or require domain labels at test time. To address this limitation, FADF leverages wavelet-based frequency statistics for intelligent test-time domain selection. After training, we compute frequency statistics for each source domain by averaging the wavelet-decomposed features across all training samples. For domain $k$, the frequency prototypes are computed as:
\begin{equation}
    \bar{F}_{\rm low}^k = \frac{1}{N_k}\sum_{i=1}^{N_k} \text{GAP}(\mathcal{W}_{\rm low}(F_i^k))
\end{equation}
\begin{equation}
    \bar{F}_{\rm high}^k = \frac{1}{N_k}\sum_{i=1}^{N_k} \text{GAP}(\mathcal{W}_{\rm high}(F_i^k))
\end{equation}
where $\text{GAP}(\cdot)$ denotes global average pooling that reduces spatial dimensions to obtain compact frequency representations $\bar{F}_{high}^k \in \mathbb{R}^{C}$. Given a test image, we extract its frequency features and compute cosine similarity with each domain's frequency prototypes:
\begin{equation}
    s_k = \frac{1}{2}\left(\frac{\bar{F}_{\rm low}^{\rm test} \cdot \bar{F}_{low}^k}{\|\bar{F}_{\rm low}^{\rm test}\| \|\bar{F}_{\rm low}^k\|} + \frac{\bar{F}_{\rm high}^{\rm test} \cdot \bar{F}_{\rm high}^k}{\|\bar{F}_{\rm high}^{\rm test}\| \|\bar{F}_{\rm high}^k\|}\right)
\end{equation}

The similarity scores are converted to fusion weights through softmax normalization:
\begin{equation}
    w_k = \frac{\exp(s_k / \tau)}{\sum_{j=1}^{K}\exp(s_j / \tau)}
\end{equation}
where $\tau$ is a temperature parameter controlling the sharpness of the weight distribution. The final domain-adaptive features are obtained by soft-weighted fusion of domain-specific outputs:
\begin{equation}
    F_{\rm fused} = \sum_{k=1}^{K} w_k \cdot F_{\rm SDM}^k
\end{equation}

This frequency-based fusion strategy enables WaveRNet to dynamically leverage multi-domain knowledge based on the frequency characteristics of test images, providing robust generalization to unseen domains without requiring explicit domain labels.

\subsection{Hierarchical Mask-Prompt Refiner}

The original SAM decoder directly upsamples low-resolution features to high-resolution masks through transposed convolutions, followed by bilinear interpolation to the target resolution. This abrupt resolution jump loses fine vessel details, particularly for tiny capillaries that require gradual reconstruction. To address this limitation, HMPR employs a coarse-to-fine refinement strategy. HMPR consists of two decoder stages operating at progressively increasing resolutions. The first decoder $\mathcal{D}_1$ generates an initial coarse mask:
\begin{equation}
    M_{256} = \mathcal{D}_1(F_{\rm fused}, P_e)
\end{equation}
where $P_e$ denotes the prompt embeddings from SAM's prompt encoder. The coarse mask $M_{256}$ is then fed back to the prompt encoder as a mask prompt, generating refined prompt embeddings that encode the predicted vessel locations. Before the second decoding stage, we apply a self-attention mechanism over the dense prompt embeddings to model long-range spatial dependencies, thereby capturing global vessel connectivity patterns that span distant image regions. The second decoder $\mathcal{D}_2$ operates with an extended upsampling path that outputs higher resolution:
\begin{equation}
    M_{512} = \mathcal{D}_2(F_{\rm fused}, \mathcal{P}(M_{256}))
\end{equation}
where $\mathcal{P}(\cdot)$ denotes the prompt encoding of the mask. The final segmentation is obtained by bilinear interpolation to the target resolution. Overall, HMPR progressively refines vessel segmentation through the mask-prompt feedback mechanism, where coarse predictions guide subsequent stages to focus on vessel regions while the attention mechanism ensures global structural consistency.

\subsection{Optimization Pipeline}

To construct the WaveRNet framework, we adopt the ViT-B architecture from SAM \cite{kirillov2023segment} as the image encoder, leveraging its powerful visual representations learned from large-scale natural image segmentation. Specifically, we load SAM's pretrained weights to initialize the image encoder and freeze most parameters to preserve the learned visual priors. To achieve parameter-efficient adaptation to the retinal imaging domain, we insert lightweight adapter layers into the transformer blocks while keeping the backbone frozen. Additionally, the wavelet transform modules $\mathcal{W}_{\rm low}$ and $\mathcal{W}_{\rm high}$, learnable domain tokens $\{t_k\}_{k=1}^{K}$ with their associated MLPs, and decoder components remain trainable to capture domain-specific frequency characteristics. During training, the gradients from the segmentation loss propagate back through HMPR to SDM, guiding the learnable wavelet transform to separate domain-invariant vessel structures from domain-specific illumination variations while simultaneously shaping domain tokens to encode unique imaging characteristics of each source domain. After training, we compute and store the frequency prototypes $\{\bar{F}_{\rm low}^k, \bar{F}_{\rm high}^k\}_{k=1}^{K}$ for each source domain, enabling efficient test-time domain adaptation through FADF. The overall training loss is formulated as:
\begin{equation}
    \mathcal{L}_{\rm total} = \lambda_1 \cdot \mathcal{L}_{\rm Dice}(\hat{y}, y) + \lambda_2 \cdot \mathcal{L}_{\rm Focal}(\hat{y}, y) + \mathcal{L}_{\rm MSE}(\hat{s}, s_{\rm IoU})
\end{equation}
where $\mathcal{L}_{\rm Dice}$ measures the volumetric overlap between predicted masks $\hat{y}$ and ground truth $y$, addressing the sparse vessel distribution. $\mathcal{L}_{\rm Focal}$ with focusing parameter $\gamma=2$ down-weights well-classified background pixels and emphasizes hard-to-segment vessel boundaries. The IoU prediction loss $\mathcal{L}_{\rm MSE}$ supervises the model's self-assessment capability, where $\hat{s}$ denotes the predicted IoU score and $s_{\rm IoU}$ represents the actual IoU. The weighting coefficients $\lambda_1$ and $\lambda_2$ are set to 1.0 and 20.0, respectively. We employ the Adam optimizer with an initial learning rate of $1 \times 10^{-4}$ and an exponential decay scheduler. The temperature parameter $\tau$ in FADF is set to 0.1 to achieve balanced fusion of domain-specific features.

\section{Experiments}
\label{sec:experiments}

\subsection{Datasets and Implementation Details}

\subsubsection{Datasets}
To comprehensively evaluate WaveRNet, we curate four publicly available retinal vessel segmentation datasets spanning diverse imaging conditions. We denote these four datasets as source domains $\mathcal{S}_1$, $\mathcal{S}_2$, $\mathcal{S}_3$, and $\mathcal{S}_4$, respectively. Notably, RECOVERY-FA19 \citep{ding2020recovery} comprises fluorescein angiography (FA) images, introducing a significant cross-modality domain shift compared to the other three color retinal datasets. The details of each dataset are as follows.

\noindent \textbf{DRIVE} \citep{staal2004drive} dataset is established for diabetic retinopathy screening and consists of 40 color retinal images (565 $\times$ 584) captured using a Canon CR5 camera. The dataset is officially divided into 20 training and 20 testing images.

\noindent \textbf{STARE} \citep{hoover2000stare} dataset comprises 20 color retinal images (700 $\times$ 605) captured using a TopCon TRV-50 camera, exhibiting diverse retinal abnormalities. We split this dataset into 15 training and 5 testing images.

\noindent \textbf{CHASE\_DB1} \citep{fraz2012chase} dataset contains 28 retinal images (999 $\times$ 960) captured from 14 multi-ethnic children using a Nidek NM-200-D handheld camera. We split this dataset into 22 training and 6 testing images.

\noindent \textbf{RECOVERY-FA19} \citep{ding2020recovery} dataset provides 8 fluorescein angiography images (3900 $\times$ 3072) captured using Optos ultra-widefield cameras, introducing cross-modality domain shift. We split this dataset into 6 training and 2 testing images.

\subsubsection{Implementation Details}

We conduct our experiments on an NVIDIA GeForce RTX 5070 Ti GPU with 16 GB memory, utilizing PyTorch 2.1.0 and CUDA 12.8. We maintain consistent training settings and configurations across all experiments to ensure fairness and reproducibility. For the optimizer, we employ Adam with a batch size of 2 and train models for 100 epochs. The initial learning rate is set to $1 \times 10^{-4}$ and is adjusted using an exponential decay scheduler with a decay factor of 0.98. The total loss function combines Dice loss and Focal loss with weights of 0.8 and 0.2, respectively. Automatic mixed precision training is enabled to accelerate computation. In our proposed WaveRNet framework, the residual weight $\alpha$ in SDM is initialized to 0.1 as a learnable parameter, the temperature $\tau$ in FADF is set to 0.5, and the domain token embeddings are initialized with a standard deviation of 0.02. All images are preprocessed following the standard SAM pipeline \citep{kirillov2023segment,ma2024segment}. The ViT-B is considered as the image encoder for all SAM-based frameworks.

For a fair comparison, we compare our WaveRNet with two categories of methods: (1) U-Net-based methods including U-Net \citep{ronneberger2015u}, UNet++ \citep{zhou2018unetpp}, UNeXt \citep{valanarasu2022unext}, Attention U-Net \citep{oktay2018attention}, DUNet \citep{jin2019dunet}, ACC-UNet \citep{ibtehaz2023accunet}, ResUNet++ \citep{jha2019resunetpp}, Swin-UNet \citep{cao2022swin}, ULite \citep{dinh20231m}, EMCADNet \citep{rahman2024emcad}, DCSAU-Net \citep{xu2023dcsau}, and MFMSNet \citep{wu2024mfmsnet}; (2) SAM-based methods including SAM \citep{kirillov2023segment}, SAM2 \citep{ravi2024sam}, MedSAM \citep{ma2024segment}, and SAM-Med2D \citep{cheng2023sammed2d}. For SAM-based methods requiring prompts, we utilize bounding box prompts automatically generated from the ground truth masks by computing the minimum enclosing rectangle.

\subsection{Evaluation Metrics and Protocols}
We evaluate the performance of all methods using three metrics: Dice coefficient (Dice), Intersection over Union (IoU), and F1-Score (F1). For all metrics, higher values indicate better performance. The best and second-best performance values are highlighted in \textbf{bold} and \underline{underlined}, respectively. We adopt two protocols for comprehensive generalization assessment:

\noindent \textbf{Intra-domain Evaluation.} The model is trained and evaluated on each dataset independently, establishing an upper-bound reference that reflects the maximum achievable performance without domain shift. This in-domain setting verifies that our frequency-guided design preserves strong discriminative capacity while pursuing cross-domain generalization.

\noindent \textbf{Leave-One-Domain-Out (LODO) Evaluation.} The model is trained on $K-1$ source domains and evaluated on the remaining unseen target domain $\mathcal{T}$. For example, when $\mathcal{T}$ = RECOVERY-FA19, the model is trained on \{DRIVE, STARE, CHASE\_DB1\}. This protocol rigorously evaluates the model's ability to generalize to completely unseen imaging conditions.

%==============================================================================
% Table 1: Comparison on Source Domains (Single-domain Training)
%==============================================================================
\subsection{Comparison on Intra-Domain Generalization}

To assess whether our domain generalization framework compromises in-domain performance, we first evaluate WaveRNet under fully supervised single-domain training, where each dataset is trained and tested independently. This experiment establishes an upper bound for segmentation accuracy and demonstrates that our frequency-guided design does not sacrifice discriminative capacity for generalization capability.

\begin{table*}[!t]
\centering
\caption{Comparison with state-of-the-art methods on in-domain evaluation. All methods are trained and tested on the same domain.}
\label{tab:source_domain}
\resizebox{\textwidth}{!}{
\begin{tabular}{l|ccc|ccc|ccc|ccc}
\toprule
\multirow{2}{*}{Methods} & \multicolumn{3}{c|}{DRIVE ($\mathcal{S}_1$)} & \multicolumn{3}{c|}{STARE ($\mathcal{S}_2$)} & \multicolumn{3}{c|}{CHASE\_DB1 ($\mathcal{S}_3$)} & \multicolumn{3}{c}{RECOVERY-FA19 ($\mathcal{S}_4$)} \\
\cmidrule{2-13}
& Dice & IoU & F1 & Dice & IoU & F1 & Dice & IoU & F1 & Dice & IoU & F1 \\
\midrule
U-Net~\citep{ronneberger2015u} & 79.54 & 66.06 & 79.88 & 70.61 & 54.70 & 72.06 & 81.13 & 68.35 & 81.27 & 56.50 & 39.38 & 56.50 \\
UNet++~\citep{zhou2018unetpp} & 79.62 & 66.17 & 79.90 & 74.12 & 58.96 & 74.83 & 81.38 & 68.73 & 81.50 & 55.72 & 38.63 & 55.72 \\
UNeXt~\citep{valanarasu2022unext} & 74.38 & 59.27 & 74.75 & 64.99 & 48.34 & 65.15 & 76.13 & 61.56 & 76.22 & 40.24 & 25.19 & 40.25 \\
Attention U-Net~\citep{oktay2018attention} & 78.68 & 64.89 & 79.01 & 71.31 & 55.46 & 71.46 & 78.09 & 64.26 & 78.38 & 54.54 & 37.50 & 54.61 \\
DUNet~\citep{jin2019dunet} & \underline{79.96} & \underline{66.64} & \underline{80.22} & 75.93 & 61.29 & 76.02 & \underline{81.89} & \underline{69.43} & \underline{81.97} & 55.50 & 38.42 & 55.52 \\
ACC-UNet~\citep{ibtehaz2023accunet} & 66.77 & 50.33 & 67.26 & 50.86 & 34.24 & 51.17 & 68.10 & 51.82 & 68.37 & 31.90 & 19.03 & 32.08 \\
ResUNet++~\citep{jha2019resunetpp} & 76.04 & 61.58 & 76.71 & 64.45 & 47.95 & 66.54 & 78.00 & 64.03 & 78.34 & 54.70 & 37.67 & 54.70 \\
Swin-UNet~\citep{cao2022swin} & 73.24 & 57.83 & 73.56 & 55.41 & 38.83 & 57.58 & 66.76 & 50.21 & 67.41 & 47.10 & 30.81 & 47.11 \\
ULite~\citep{dinh20231m} & 74.81 & 59.80 & 75.06 & 66.19 & 49.58 & 67.02 & 74.22 & 59.08 & 74.35 & 39.53 & 24.67 & 39.70 \\
EMCADNet~\citep{rahman2024emcad} & 73.14 & 57.70 & 73.72 & 64.05 & 47.55 & 67.20 & 73.86 & 58.64 & 74.07 & 44.82 & 28.89 & 44.83 \\
DCSAU-Net~\citep{xu2023dcsau} & 78.18 & 64.23 & 78.56 & 69.78 & 53.63 & 70.22 & 78.45 & 64.64 & 78.59 & 54.17 & 37.21 & 54.21 \\
MFMSNet~\citep{wu2024mfmsnet} & 77.43 & 63.22 & 77.82 & 64.93 & 48.12 & 65.73 & 77.66 & 63.62 & 77.82 & 42.13 & 26.69 & 42.16 \\
\midrule
SAM-FT~\citep{kirillov2023segment} & 69.67 & 53.48 & 83.43 & 70.78 & 54.89 & 84.31 & 69.20 & 52.99 & 83.55 & 38.19 & 23.62 & 66.48 \\
MedSAM-FT~\citep{ma2024segment} & 41.06 & 26.53 & 68.66 & 16.76 & 10.65 & 56.49 & 23.59 & 13.44 & 60.07 & 0.00 & 0.00 & 47.04 \\
\midrule
WaveRNet (Ours) & \textbf{80.46} & \textbf{67.32} & \textbf{80.50} & \textbf{79.39} & \textbf{65.91} & \textbf{79.40} & \textbf{81.94} & \textbf{69.48} & \textbf{82.13} & \textbf{60.33} & \textbf{43.22} & \textbf{60.25} \\
\bottomrule
\end{tabular}
}
\end{table*}

\begin{figure*}[!t]
\centering
\includegraphics[width=0.9\textwidth]{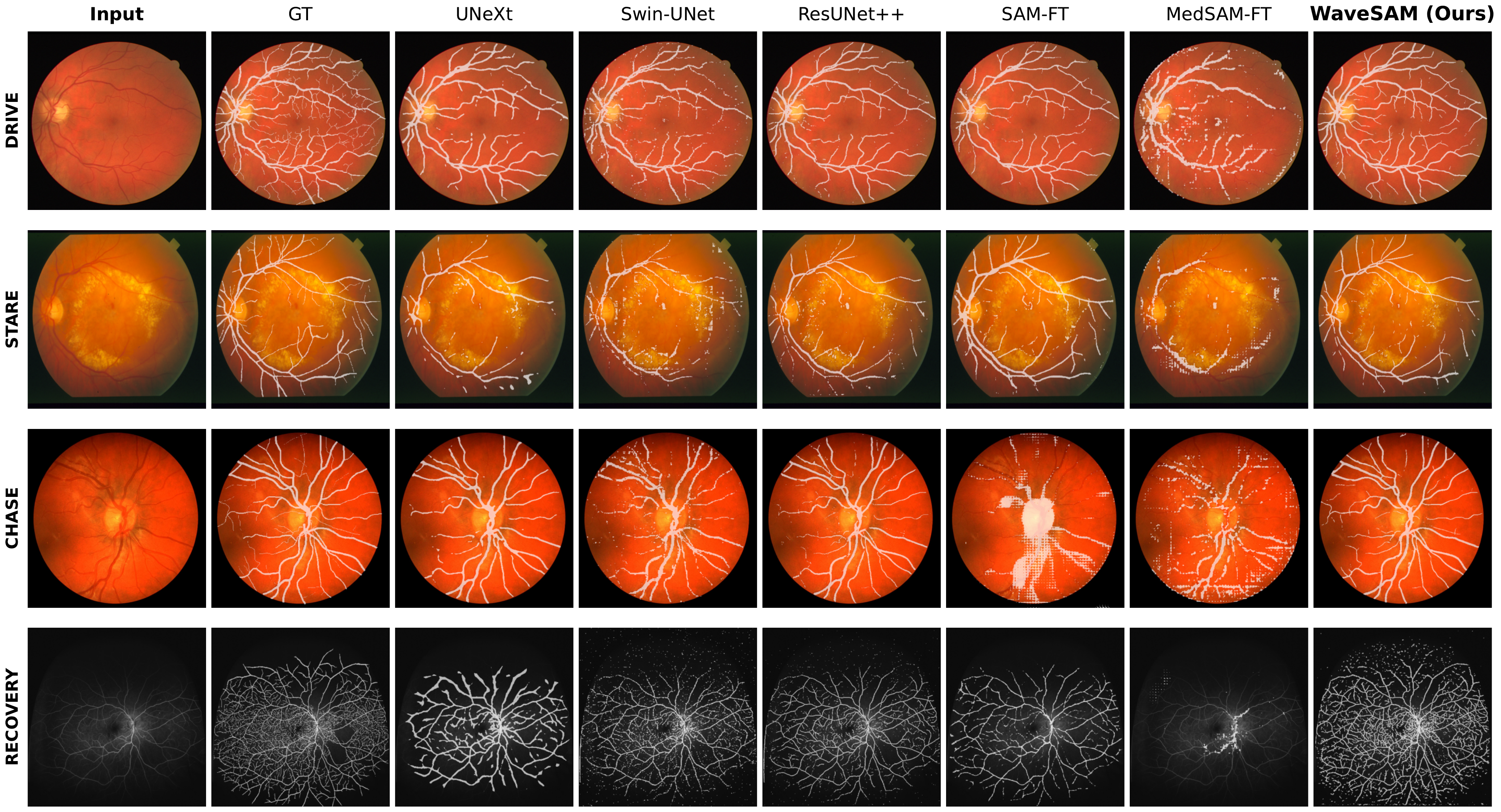}
\caption{Qualitative comparison of segmentation results under single-domain training. From left to right: input image, ground truth, UNeXt, Swin-UNet, ResUNet++, SAM-FT, MedSAM-FT, and WaveRNet (Ours). Each row represents a different dataset: DRIVE ($\mathcal{S}_1$), STARE ($\mathcal{S}_2$), CHASE\_DB1 ($\mathcal{S}_3$), and RECOVERY-FA19 ($\mathcal{S}_4$). White overlays indicate the predicted vessel segmentation masks.}
\label{fig:visualization_single}
\end{figure*}

As shown in Table~\ref{tab:source_domain}, WaveRNet achieves the best performance across all four datasets. On DRIVE and CHASE\_DB1, WaveRNet outperforms the second-best method DUNet by 0.50\% and 0.05\% in Dice, respectively. Notably, on STARE, WaveRNet achieves 79.39\% Dice, surpassing DUNet by a substantial margin of 3.46\%. The most significant improvement is observed on RECOVERY-FA19, where WaveRNet achieves 60.33\% Dice compared to 56.50\% for U-Net, representing a 3.83\% absolute gain. This dataset presents unique challenges due to its fluorescein angiography modality with distinct imaging characteristics. The consistent improvements across datasets with varying imaging conditions validate that the frequency-aware feature decomposition in SDM enhances discriminative capacity rather than compromising it. These results demonstrate that WaveRNet not only excels in cross-domain scenarios but also maintains competitive performance when training and testing data share the same distribution.

Fig.~\ref{fig:visualization_single} presents qualitative comparisons under single-domain training. WaveRNet consistently produces segmentation masks that closely match the ground truth across all four datasets. Compared to U-Net-based methods (UNeXt, Swin-UNet, ResUNet++), WaveRNet captures finer vessel branches and maintains better connectivity. Notably, SAM-FT and MedSAM-FT exhibit severe over-segmentation artifacts, particularly visible in CHASE where MedSAM-FT produces extensive false positives around the optic disc region. In contrast, WaveRNet achieves clean segmentation boundaries with minimal noise, demonstrating the effectiveness of our frequency-guided approach for in-domain vessel segmentation.

\begin{table*}[!t]
\centering
\caption{Comparison with state-of-the-art methods under the LODO protocol. The model is trained on $K-1$ source domains and evaluated on the remaining unseen target domain $\mathcal{T}$.}
\label{tab:target_domain}
\resizebox{\textwidth}{!}{
\begin{tabular}{l|ccc|ccc|ccc|ccc|c}
\toprule
\multirow{2}{*}{Methods} & \multicolumn{3}{c|}{$\mathcal{T}$ = DRIVE} & \multicolumn{3}{c|}{$\mathcal{T}$ = STARE} & \multicolumn{3}{c|}{$\mathcal{T}$ = CHASE\_DB1} & \multicolumn{3}{c|}{$\mathcal{T}$ = RECOVERY-FA19} & \multirow{2}{*}{Avg.} \\
\cmidrule{2-13}
& Dice & IoU & F1 & Dice & IoU & F1 & Dice & IoU & F1 & Dice & IoU & F1 & \\
\midrule
U-Net~\citep{ronneberger2015u} & 44.59 & 28.73 & 63.94 & 39.19 & 25.05 & 62.61 & 37.84 & 23.37 & 61.34 & 8.91 & 4.67 & 0.24 & 32.63 \\
UNet++~\citep{zhou2018unetpp} & 30.30 & 17.89 & 63.69 & 33.32 & 20.45 & 61.37 & 25.98 & 14.95 & 59.85 & 12.73 & 6.80 & 1.51 & 25.58 \\
UNeXt~\citep{valanarasu2022unext} & 14.54 & 7.85 & 3.53 & 12.65 & 6.77 & 0.01 & 11.48 & 6.09 & 0.00 & 15.75 & 8.56 & 16.90 & 13.61 \\
Attention U-Net~\citep{oktay2018attention} & 18.40 & 10.14 & 33.39 & 25.73 & 14.93 & 60.20 & 18.18 & 10.01 & 46.80 & 16.04 & 8.73 & 22.89 & 19.59 \\
DUNet~\citep{jin2019dunet} & 26.28 & 15.15 & 63.05 & 22.26 & 12.65 & 60.43 & 21.18 & 11.87 & 60.58 & 13.86 & 7.45 & 0.04 & 20.90 \\
ACC-UNet~\citep{ibtehaz2023accunet} & 14.95 & 8.09 & 15.67 & 14.05 & 7.57 & 17.84 & 12.71 & 6.80 & 13.52 & 15.43 & 8.37 & 13.12 & 14.29 \\
ResUNet++~\citep{jha2019resunetpp} & 29.46 & 17.32 & 60.34 & 33.25 & 20.32 & 57.17 & 38.02 & 23.51 & 59.96 & 12.41 & 6.62 & 0.46 & 28.29 \\
ULite~\citep{dinh20231m} & 34.47 & 20.86 & 52.04 & 18.27 & 10.09 & 31.44 & 13.61 & 7.31 & 22.82 & 13.21 & 7.08 & 1.10 & 19.89 \\
EMCADNet~\citep{rahman2024emcad} & 15.45 & 8.38 & 15.99 & 13.42 & 7.21 & 14.14 & 12.22 & 6.51 & 12.97 & 15.96 & 8.68 & 17.08 & 14.26 \\
DCSAU-Net~\citep{xu2023dcsau} & 28.20 & 16.45 & 62.46 & 26.12 & 15.24 & 57.68 & 23.70 & 13.48 & 61.59 & 11.93 & 6.35 & 2.21 & 22.49 \\
MFMSNet~\citep{wu2024mfmsnet} & \underline{59.28} & \underline{42.20} & \underline{65.33} & 48.95 & 34.00 & 58.05 & \underline{47.52} & \underline{31.22} & 55.80 & 7.02 & 3.65 & 4.85 & 40.69 \\
Swin-UNet~\citep{cao2022swin} & 49.40 & 32.91 & 59.85 & \underline{50.43} & \underline{35.29} & \underline{59.87} & 45.06 & 29.21 & \underline{57.52} & 4.72 & 2.42 & 0.01 & 37.40 \\
\midrule
SAM-FT~\citep{kirillov2023segment} & 72.49 & 56.90 & 72.74 & 70.96 & 55.05 & 71.45 & 42.38 & 26.92 & 43.02 & 36.35 & 22.33 & 36.09 & 55.55 \\
SAM2-FT~\citep{ravi2024sam} & 66.64 & 50.08 & 66.90 & 45.74 & 31.94 & 49.94 & 65.31 & 48.73 & 65.70 & 31.92 & 19.00 & 36.97 & 52.40 \\
MedSAM-FT~\citep{ma2024segment} & 51.84 & 35.17 & 52.18 & 31.42 & 19.91 & 35.25 & 36.21 & 22.33 & 37.11 & 6.10 & 3.16 & 6.34 & 31.39 \\
SAM-Med2D-FT~\citep{cheng2023sammed2d} & 69.72 & 53.53 & 69.95 & 66.28 & 49.61 & 66.65 & 65.09 & 48.42 & 65.31 & \underline{39.18} & \underline{24.39} & \underline{39.22} & \underline{60.07} \\
\midrule
WaveRNet (Ours) & \textbf{78.55} & \textbf{64.71} & \textbf{78.59} & \textbf{81.06} & \textbf{68.29} & \textbf{81.45} & \textbf{76.58} & \textbf{62.12} & \textbf{76.75} & \textbf{41.75} & \textbf{26.42} & \textbf{41.66} & \textbf{69.49} \\
\bottomrule
\end{tabular}
}
\end{table*}

\begin{figure*}[!t]
\centering
\includegraphics[width=0.9\textwidth]{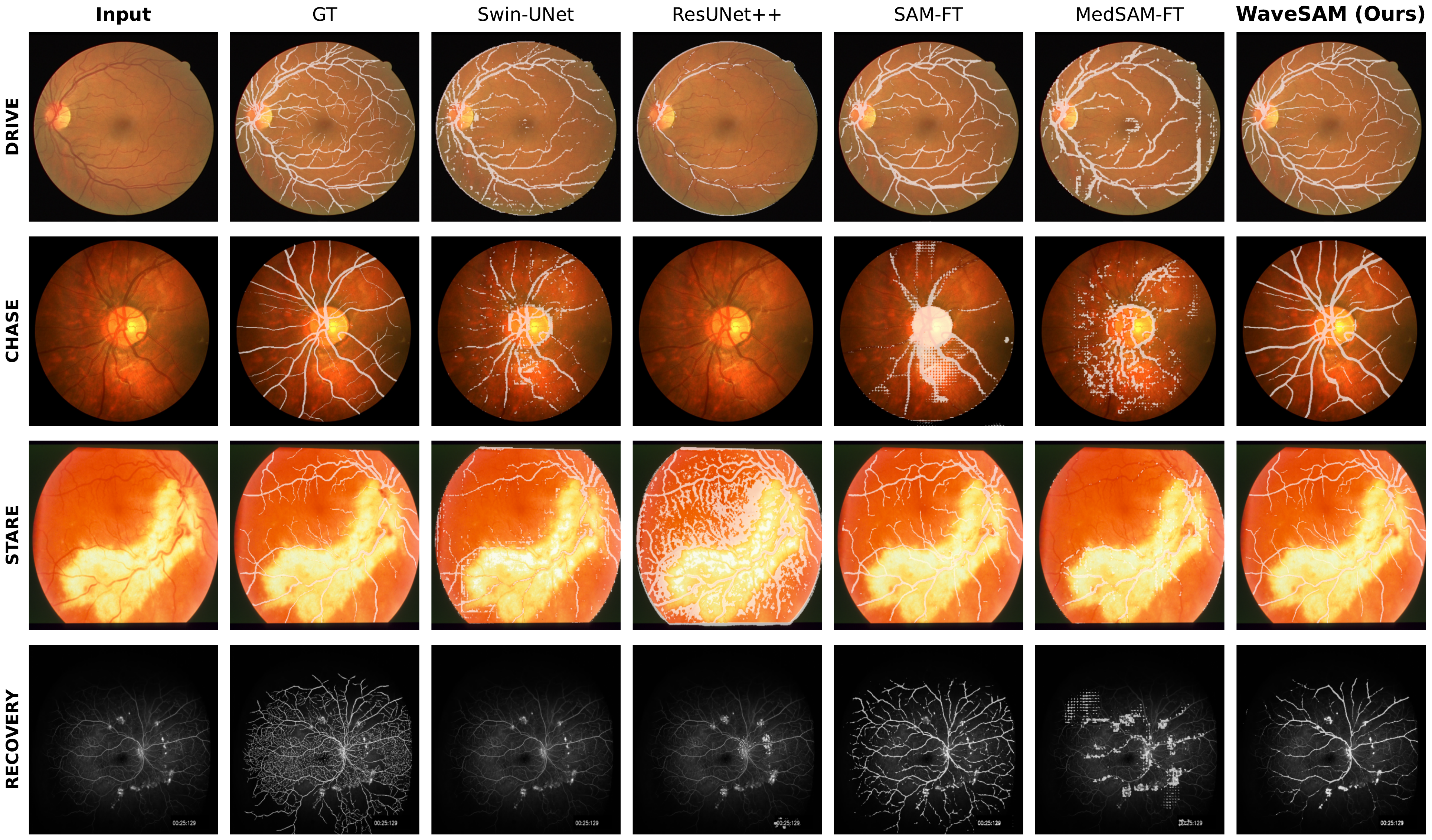}
\caption{Qualitative comparison of segmentation results on unseen target domains under the LODO protocol. From left to right: input image, ground truth, UNet++, Swin-UNet, ResUNet++, SAM-FT, MedSAM-FT, and WaveRNet (Ours). Each row represents a different unseen target domain: DRIVE ($\mathcal{T}=\mathcal{S}_1$), CHASE\_DB1 ($\mathcal{T}=\mathcal{S}_3$), STARE ($\mathcal{T}=\mathcal{S}_2$), and RECOVERY-FA19 ($\mathcal{T}=\mathcal{S}_4$). White overlays indicate the predicted vessel segmentation masks.}
\label{fig:visualization}
\end{figure*}

%==============================================================================
% Table 2: Comparison on Target Domains - Traditional Methods (LODO)
%==============================================================================
\begin{table*}[!t]
\centering
\caption{Ablation study of WaveRNet components under mixed domain training. $M_1$: SDM. $M_2$: FADF. $M_3$: HMPR.}
\label{tab:ablation}
\resizebox{\textwidth}{!}{
\begin{tabular}{ccc|ccc|ccc|ccc|ccc|ccc}
\toprule
\multirow{2}{*}{$M_1$} & \multirow{2}{*}{$M_2$} & \multirow{2}{*}{$M_3$} & \multicolumn{3}{c|}{DRIVE ($\mathcal{S}_1$)} & \multicolumn{3}{c|}{STARE ($\mathcal{S}_2$)} & \multicolumn{3}{c|}{CHASE\_DB1 ($\mathcal{S}_3$)} & \multicolumn{3}{c|}{RECOVERY-FA19 ($\mathcal{S}_4$)} & \multicolumn{3}{c}{Average} \\
\cmidrule{4-18}
& & & Dice & IoU & F1 & Dice & IoU & F1 & Dice & IoU & F1 & Dice & IoU & F1 & Dice & IoU & F1 \\
\midrule
& & & 76.00 & 61.32 & 76.00 & 76.11 & 61.51 & 76.11 & 79.34 & 65.81 & 79.34 & 52.90 & 35.99 & 52.90 & 71.09 & 56.16 & 71.09 \\
$\checkmark$ & & & 76.63 & 62.15 & 76.63 & 76.20 & 61.67 & 76.20 & 80.66 & 67.66 & 80.66 & 54.86 & 37.82 & 54.86 & 72.09 & 57.33 & 72.09 \\
& $\checkmark$ & & 76.47 & 61.93 & 76.47 & 76.74 & 62.39 & 76.74 & 80.31 & 67.18 & 80.31 & 53.13 & 36.22 & 53.13 & 71.66 & 56.93 & 71.66 \\
& & $\checkmark$ & 78.06 & 64.04 & 78.06 & 77.16 & 62.92 & 77.16 & 80.57 & 67.52 & 80.57 & 57.92 & 40.81 & 57.92 & 73.43 & 58.82 & 73.43 \\
$\checkmark$ & $\checkmark$ & & 78.22 & 64.26 & 78.22 & 76.96 & 62.68 & 76.96 & 80.96 & 68.07 & 80.96 & 58.55 & 41.43 & 58.55 & 73.67 & 59.11 & 73.67 \\
$\checkmark$ & & $\checkmark$ & 80.18 & 66.93 & 80.22 & \textbf{79.61} & \textbf{66.25} & 79.40 & 80.14 & 67.02 & 80.28 & 58.82 & 41.67 & 58.76 & 74.69 & 60.47 & 74.68 \\
& $\checkmark$ & $\checkmark$ & 78.32 & 64.38 & 78.32 & 76.99 & 62.72 & 76.99 & 80.79 & 67.85 & 80.79 & 58.41 & 41.29 & 58.41 & 73.63 & 59.06 & 73.63 \\
$\checkmark$ & $\checkmark$ & $\checkmark$ & \textbf{80.46} & \textbf{67.32} & \textbf{80.50} & 79.39 & 65.91 & \textbf{79.46} & \textbf{81.94} & \textbf{69.48} & \textbf{82.13} & \textbf{60.33} & \textbf{43.22} & \textbf{60.25} & \textbf{75.53} & \textbf{61.48} & \textbf{75.57} \\
\bottomrule
\end{tabular}
}
\end{table*}

\begin{table*}[!t]
\centering
\caption{Ablation study of WaveRNet components under the LODO protocol. $M_1$: SDM. $M_2$: FADF. $M_3$: HMPR.}
\label{tab:ablation_lodo}
\resizebox{\textwidth}{!}{
\begin{tabular}{ccc|ccc|ccc|ccc|ccc|ccc}
\toprule
\multirow{2}{*}{$M_1$} & \multirow{2}{*}{$M_2$} & \multirow{2}{*}{$M_3$} & \multicolumn{3}{c|}{DRIVE ($\mathcal{S}_1$)} & \multicolumn{3}{c|}{STARE ($\mathcal{S}_2$)} & \multicolumn{3}{c|}{CHASE\_DB1 ($\mathcal{S}_3$)} & \multicolumn{3}{c|}{RECOVERY-FA19 ($\mathcal{S}_4$)} & \multicolumn{3}{c}{Average} \\
\cmidrule{4-18}
& & & Dice & IoU & F1 & Dice & IoU & F1 & Dice & IoU & F1 & Dice & IoU & F1 & Dice & IoU & F1 \\
\midrule
 &  &  & 75.34 & 60.49 & 75.95 & 76.34 & 61.91 & 77.25 & 75.46 & 60.67 & 75.90 & 13.94 & 7.50 & 0.00 & 60.27 & 47.64 & 57.28 \\
$\checkmark$ &  &  & 76.80 & 62.40 & 77.09 & 79.47 & 66.09 & 80.14 & 76.62 & 62.17 & 77.03 & 35.36 & 21.57 & 35.43 & 67.06 & 53.06 & 67.42 \\
 & $\checkmark$ &  & 76.55 & 62.07 & 76.89 & 78.93 & 65.32 & 79.65 & 76.62 & 62.17 & 77.03 & 14.64 & 7.95 & 14.48 & 61.69 & 49.38 & 62.01 \\
 &  & $\checkmark$ & 76.47 & 61.96 & 66.01 & 26.14 & 15.06 & 19.55 & 74.91 & 60.05 & 62.25 & 22.56 & 12.77 & 11.77 & 50.02 & 37.46 & 39.90 \\
$\checkmark$ & $\checkmark$ &  & \textbf{78.91} & \textbf{65.20} & 77.81 & 74.66 & 59.65 & 80.91 & 76.51 & 62.13 & 74.98 & 36.93 & 22.74 & 32.27 & 66.75 & 52.43 & 66.49 \\
$\checkmark$ &  & $\checkmark$ & 77.53 & 63.37 & 77.80 & 79.63 & 66.28 & 80.31 & \textbf{76.96} & \textbf{62.61} & \textbf{77.39} & 31.97 & 19.09 & 32.15 & 66.52 & 52.84 & 66.91 \\
 & $\checkmark$ & $\checkmark$ & 76.98 & 62.63 & 77.24 & 79.47 & 66.09 & 80.14 & 76.89 & 62.52 & 77.27 & 29.96 & 17.64 & 30.13 & 65.83 & 52.22 & 66.20 \\
$\checkmark$ & $\checkmark$ & $\checkmark$ & 78.55 & 64.71 & \textbf{78.59} & \textbf{81.06} & \textbf{68.29} & \textbf{81.45} & 76.58 & 62.12 & 76.75 & \textbf{41.75} & \textbf{26.42} & \textbf{41.66} & \textbf{69.49} & \textbf{55.39} & \textbf{69.61} \\
\bottomrule
\end{tabular}
}
\end{table*}

\begin{table}[!t]
\centering
\caption{Ablation study of SDM frequency branches under the LODO protocol. $F_{\rm low}$: low-frequency branch. $F_{\rm high}$: high-frequency branch.}
\label{tab:wavelet_analysis}
\small
\setlength\tabcolsep{4pt}
\begin{tabular}{cc|cccc|c}
\toprule
$F_{\rm low}$ & $F_{\rm high}$ & DRIVE & STARE & CHASE & RECOVERY & Avg. \\
\midrule
 &  & 54.61 & 37.15 & 54.90 & 0.44 & 36.77 \\
$\checkmark$ &  & 54.90 & \textbf{55.56} & 45.24 & \textbf{1.26} & 39.24 \\
 & $\checkmark$ & 55.31 & 53.64 & 54.07 & 0.29 & 40.83 \\
$\checkmark$ & $\checkmark$ & \textbf{55.33} & 55.26 & \textbf{56.68} & 0.73 & \textbf{42.00} \\
\bottomrule
\end{tabular}
\end{table}

\subsection{Comparison on Multi-Source Domain Generalization}

We employ a standard leave-one-domain-out (LODO) strategy to evaluate the domain generalization capability. Specifically, the model is trained on $K-1$ source domains and evaluated on the remaining unseen target domain. For example, when testing on RECOVERY-FA19 ($\mathcal{S}_4$), the model is trained on $\{\mathcal{S}_1, \mathcal{S}_2, \mathcal{S}_3\}$ (DRIVE, STARE, CHASE\_DB1). This protocol rigorously evaluates the model's ability to generalize to completely unseen imaging conditions. Table~\ref{tab:target_domain} presents the domain generalization comparison under the LODO protocol. Several noteworthy observations emerge from these results. First, conventional U-Net-based segmentation networks suffer catastrophic performance degradation when evaluated on unseen domains. For instance, DUNet achieves 79.96\% Dice on DRIVE under single-domain training (Table~\ref{tab:source_domain}), but drops dramatically to 26.28\% under the LODO protocol, representing a 53.68\% absolute decrease. This phenomenon highlights the inherent limitation of all U-Net variants in handling domain shift. In addition, while SAM-based methods generally outperform U-Net variants in domain generalization, simply fine-tuning SAM on medical data is insufficient for optimal cross-domain performance. MedSAM-FT, despite being trained on large-scale medical datasets, achieves only 31.39\% average Dice, performing worse than the vanilla SAM-FT (55.55\%). This suggests that extensive medical fine-tuning without domain-aware mechanisms can actually impair generalization. In stark contrast, WaveRNet exhibits exceptional robustness, achieving an average Dice of 69.49\% that surpasses the second-best SAM-based method SAM-Med2D-FT (60.07\%) by 9.42\% and the leading U-Net-based method MFMSNet (40.69\%) by 28.80\%. The improvements are particularly pronounced on CHASE\_DB1 (+11.49\% over SAM-Med2D-FT), DRIVE (+8.83\%), and STARE (+14.78\%). Even on RECOVERY-FA19, which presents extreme cross-modality challenges due to its fluorescein angiography imaging, WaveRNet achieves 41.75\% Dice while most U-Net variants fail completely (below 16\%). These results validate that our frequency-guided domain modulation effectively complements SAM's pretrained representations for robust domain generalization.

Fig.~\ref{fig:visualization} presents qualitative comparisons on representative samples from each unseen target domain under the LODO protocol. Specifically, U-Net-based methods (UNet++, Swin-UNet, ResUNet++) exhibit severe under-segmentation on unseen domains, missing large portions of the vessel tree. This aligns with their catastrophic Dice drops observed in Table~\ref{tab:target_domain}. Second, SAM-based methods (SAM-FT, MedSAM-FT) produce more complete vessel structures but suffer from excessive false positives, particularly visible in CHASE\_DB1 and STARE where they generate noisy predictions around the optic disc region. Third, WaveRNet achieves the most accurate segmentation with clear vessel boundaries, complete branching structures, and minimal false positives. The improvements are especially visible on RECOVERY-FA19 (fluorescein angiography), where most methods fail to capture the fine vessel details while WaveRNet successfully preserves the complete vessel tree structure, validating the effectiveness of our frequency-guided domain adaptation.

%==============================================================================
% Table 3: Ablation Study (Mixed Domain Training)
%==============================================================================
\subsection{Ablation Study}
To investigate the contribution of each proposed component, we conduct comprehensive ablation experiments under mixed domain training, where all four domains are used for both training and testing. We systematically evaluate eight configurations by progressively enabling the Spectral-guided Domain Modulator (SDM), Frequency-Adaptive Domain Fusion (FADF), and Hierarchical Mask-Prompt Refiner (HMPR). Table~\ref{tab:ablation} presents the ablation results. Firstly, adding SDM to the baseline improves the average Dice from 71.09\% to 72.09\% (+1.00\%). The improvement validates that wavelet-based frequency decomposition effectively captures domain-invariant features by separating illumination-stable low-frequency structures from contrast-sensitive high-frequency boundaries. FADF alone brings a 0.57\% improvement in average Dice (71.09\% $\rightarrow$ 71.66\%). Combining SDM and FADF yields 73.67\% average Dice, demonstrating synergistic effects (+2.58\% over baseline). This synergy arises because SDM generates frequency-aware domain-specific features that FADF can effectively leverage for domain adaptation. HMPR contributes a 2.34\% improvement when added independently (71.09\% $\rightarrow$ 73.43\%), representing the largest single-module gain. This confirms that progressive coarse-to-fine refinement with mask-prompt feedback effectively recovers fine vessel structures lost in SAM's direct upsampling. The full WaveRNet (SDM + FADF + HMPR) achieves 75.53\% average Dice, outperforming all partial configurations. Compared to the best two-component combination (SDM + HMPR: 74.69\%), the full model gains an additional 0.84\%, indicating that all three modules complement each other for optimal performance. 

To further validate the contribution of each component under domain shift conditions, we conduct additional ablation experiments using the LODO protocol, as shown in Table~\ref{tab:ablation_lodo}. Under the LODO protocol, SDM demonstrates its critical role in domain generalization, improving average Dice from 60.27\% to 67.06\% (+6.79\%). This substantial gain highlights that frequency-aware feature decomposition is particularly effective when facing unseen domains. Notably, HMPR alone shows degraded performance (50.02\%), suggesting that hierarchical refinement requires frequency-guided features to function optimally under domain shift. The full model achieves 69.49\% average Dice, with the most significant improvement on RECOVERY-FA19 (+27.81\% over baseline), demonstrating WaveRNet's robustness to cross-modality domain shift.

%==============================================================================
% Analysis of Wavelet Decomposition
%==============================================================================
\subsubsection{Analysis of Frequency Decomposition}
To validate our hypothesis that both high-frequency and low-frequency components are essential for robust domain generalization, we conduct a detailed ablation study on the wavelet decomposition branches in SDM. As shown in Table~\ref{tab:wavelet_analysis}, we systematically evaluate four configurations: (1) baseline without frequency decomposition, (2) low-frequency branch only, (3) high-frequency branch only, and (4) full dual-branch design. Several key observations emerge from the results. First, incorporating either frequency branch alone yields improvements over the baseline ($F_{\rm low}$: +2.47\%, $F_{\rm high}$: +4.06\%), demonstrating that explicit frequency-aware modeling benefits cross-domain generalization. Second, combining both branches achieves the highest average Dice score of 42.00\% (+5.23\% over baseline), confirming that low-frequency and high-frequency components provide complementary information. The low-frequency branch captures illumination-robust global structures that remain stable across imaging devices, while the high-frequency branch preserves contrast-sensitive vessel boundaries critical for fine-grained segmentation. Third, examining individual domains reveals that $F_{\rm low}$ contributes more significantly to STARE and RECOVERY-FA19, which exhibit pronounced illumination variations, whereas $F_{\rm high}$ shows stronger gains on DRIVE and CHASE\_DB1, where contrast differences dominate. By explicitly decoupling and modeling both frequency bands, SDM effectively addresses the heterogeneous domain shift characteristics inherent in multi-source retinal imaging. Future work will explore unsupervised domain discovery mechanisms to eliminate the need for explicit domain labels, investigate efficient decoder architectures for reduced inference latency, and extend validation to broader ophthalmic imaging modalities.

%==============================================================================
% Discussion
%==============================================================================
% \subsection{Discussion}

% While WaveRNet demonstrates strong cross-domain generalization, several limitations warrant discussion. First, the current framework requires domain labels during training to learn domain-specific tokens, which may limit scalability to scenarios with numerous unlabeled source domains. Second, the hierarchical refinement in HMPR introduces additional computational overhead compared to single-stage decoders, potentially affecting real-time clinical deployment. Third, although RECOVERY-FA19 introduces cross-modality evaluation, the framework has not been extensively validated on other imaging modalities such as optical coherence tomography (OCT) or ultra-widefield imaging. Future work will explore unsupervised domain discovery mechanisms to eliminate the need for explicit domain labels, investigate efficient decoder architectures for reduced inference latency, and extend validation to broader ophthalmic imaging modalities.

\section{Conclusion}
\label{sec:conclusion}

In this paper, we have proposed WaveRNet, a wavelet-guided framework for robust domain-generalized retinal vessel segmentation. The proposed SDM integrates wavelet transform with domain-specific modulation, effectively decomposing features into high-frequency and low-frequency components to address illumination and contrast variations across different imaging domains. The FADF enables intelligent test-time domain selection via wavelet-based frequency similarity computation and soft-weighted fusion across source domains. The HMPR employs a coarse-to-fine mask-prompt refinement strategy with long-range dependency modeling to progressively recover fine vessel structures. Extensive experiments under the Leave-One-Domain-Out protocol on four public retinal datasets demonstrate that WaveRNet achieves state-of-the-art performance with superior cross-domain robustness. The strong generalization performance suggests that WaveRNet could facilitate the deployment of automated vessel analysis systems across heterogeneous clinical environments without site-specific retraining, potentially accelerating the translation of AI-assisted ophthalmic diagnosis into routine clinical practice.

\section*{Declaration of competing interest}

The authors declare that they have no known competing financial interests or personal relationships that could have appeared to influence the work reported in this paper.

\section*{Data Availability}

All the data used in this study are obtained from publicly accessible datasets, and the code will be made available on GitHub upon publication.

% TODO: Add acknowledgments if any

%% Harvard style references
\bibliographystyle{model2-names.bst}\biboptions{authoryear}
\bibliography{refs}

\end{document}